\pdfoutput=1

\documentclass[11pt]{article}

\usepackage{authblk}

\usepackage{times}
\usepackage{latexsym}
\usepackage{microtype}
\usepackage{mathtools}
\usepackage{algorithm}
\usepackage[noend]{algorithmic}

\usepackage[]{ACL2023}

\usepackage{amsmath,amsfonts,bm}

\def\eqref#1{equation~\ref{#1}}

\def\1{\bm{1}}

\DeclareMathAlphabet{\mathsfit}{\encodingdefault}{\sfdefault}{m}{sl}
\SetMathAlphabet{\mathsfit}{bold}{\encodingdefault}{\sfdefault}{bx}{n}

\newcommand{\E}{\mathbb{E}}

\usepackage{hyperref}
\usepackage{url}
\usepackage{graphicx}
\usepackage{comment}
\usepackage{xspace}
\usepackage{booktabs}
\usepackage{xcolor}
\usepackage{microtype}
\usepackage{inconsolata}

\newcommand*{\img}[1]{%
    \raisebox{-.15\baselineskip}{%
        \includegraphics[
        height=\baselineskip,
        width=\baselineskip,
        keepaspectratio,
        ]{#1}%
    }%
}

\title{\img{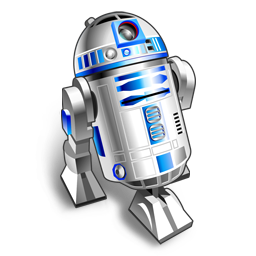}I2D2: Inductive Knowledge Distillation with\\ NeuroLogic and Self-Imitation}

\author{\parbox{\linewidth}{\centering
Chandra Bhagavatula$^*$, 
Jena D. Hwang$^*$, 
Doug Downey$^\mathparagraph{^*}$, 
Ronan Le Bras$^*$,\\
Ximing Lu$^\mathsection{^*}$, 
Lianhui Qin$^\mathsection$,
Keisuke Sakaguchi$^\ddagger$, 
Swabha Swayamdipta$^\dagger$, 
Peter West$^\mathsection{^*}$,\\
Yejin Choi$^\mathsection{^*}$
}
}
\affil{$^*$Allen Institute for AI, $^\dagger$University of Southern California, \\$^\ddagger$Tohoku University, $^\mathparagraph$Northwestern University, $^\mathsection$University of Washington\\\href{https://i2d2.allen.ai/}{\color{purple}{i2d2.allen.ai}}}

\newcommand{\corpus}{\texttt{Gen-A-tomic}\xspace}

\newcommand{\framework}{{{\textsc{I2D2}}}\xspace}

\newcommand{\gptthree}{GPT-3\xspace}
\newcommand{\gpttwo}{GPT-2\xspace}
\newcommand{\gpttwoxl}{GPT-2 XL\xspace}

\begin{document}

\maketitle

\begin{abstract}

Commonsense capabilities of pre-trained language models dramatically improve with scale, leading many to believe that scale is the only winning recipe. But is it? Here, we investigate an alternative that \textit{a priori} seems impossible: 
can smaller language models (e.g., GPT-2) win over models that are orders of magnitude larger and better (e.g., GPT-3), if powered with novel commonsense distillation algorithms?
The key intellectual challenge is to design a learning algorithm that achieves a competitive level of commonsense acquisition, without relying on the benefits of scale. In particular, we study \emph{generative} models of commonsense knowledge, focusing on the task of generating \emph{generics}, statements of commonsense facts about everyday concepts, e.g., birds can fly.

We introduce \framework{}, a novel commonsense distillation framework that loosely follows \citet{west-etal-2022-symbolic}'s Symbolic Knowledge Distillation but breaks the dependence on the extreme-scale teacher model with two innovations: (1) the novel adaptation of \textbf{NeuroLogic} Decoding \cite{lu2020neurologic} to enhance the generation quality of the weak, off-the-shelf language models, and (2) \textbf{self-imitation learning} to iteratively learn from the model's own enhanced commonsense acquisition capabilities. Empirical results suggest that scale is not the only way, as novel algorithms can be a promising alternative. Moreover, our study leads to a new corpus of generics, \corpus, that is the largest and highest-quality available to date.

\end{abstract}

\begin{figure}[ht!]
\centering
  \includegraphics[width=.45\textwidth]{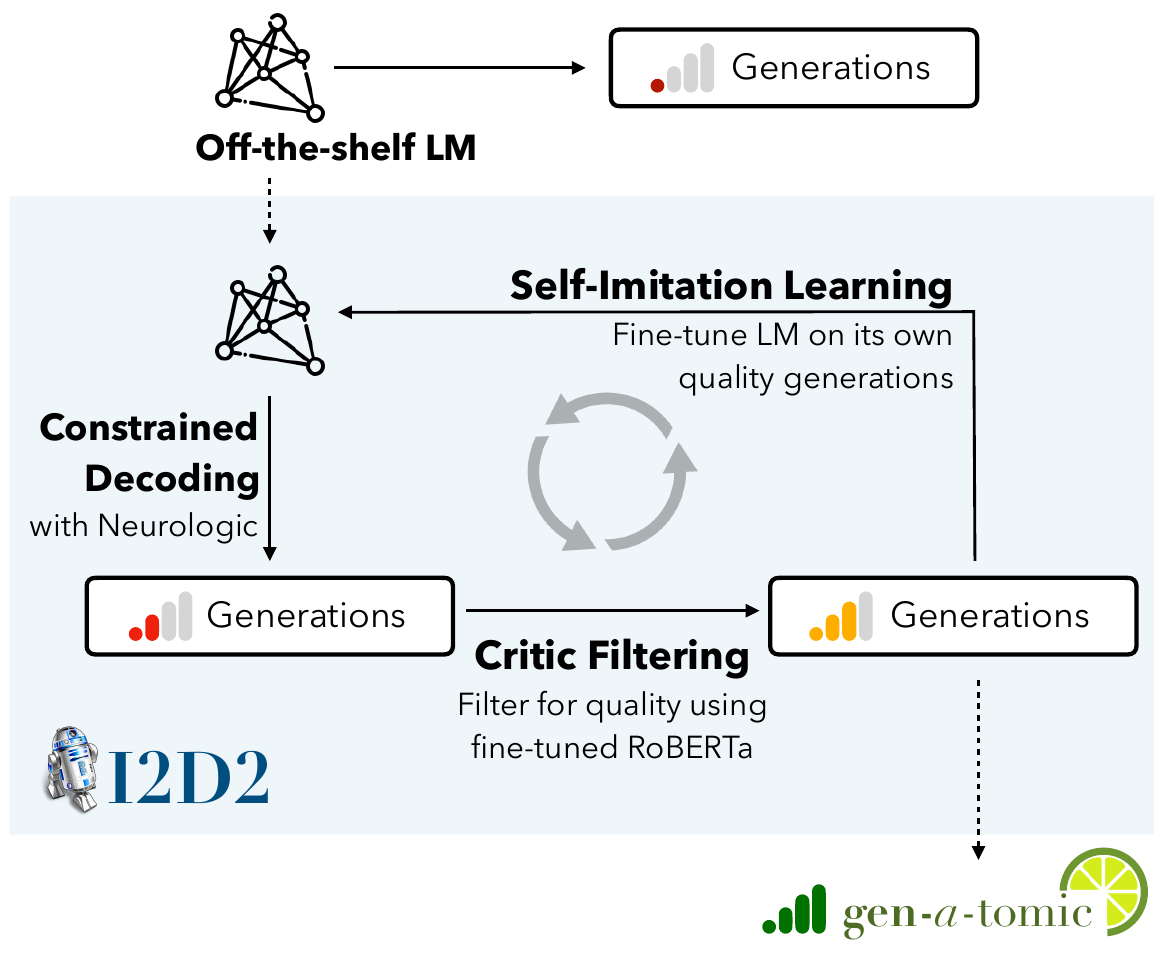} %
  \caption{\framework radically improves over off-the-shelf generation from \gpttwoxl using constrained decoding and self-imitation. }
  \label{fig:fig1}
\end{figure}

\section{Introduction}

Language models (LMs) become better with scale.
However, even the largest LMs continue to fail in unexpected ways due to their lack of commonsense \citep{brachman2022}. \textit{Knowledge models} -- custom LMs trained to generate knowledge---provide on-demand access to task-specific knowledge to address this gap \cite{bosselut2019comet}. 
Today, the best strategy for training a knowledge model depends on large-scale, albeit noisy knowledge generated from a large LM \cite{west-etal-2022-symbolic}. 
Are massive-scale LMs the only way to build commonsense capabilities?
In addition to being an interesting scientific inquiry, if smaller LMs can indeed generate high-quality commonsense, training knowledge models will become far more efficient and accessible compared to the state-of-the-art.

\begin{figure*}[t!]
\centering
  \includegraphics[width=0.99\textwidth]{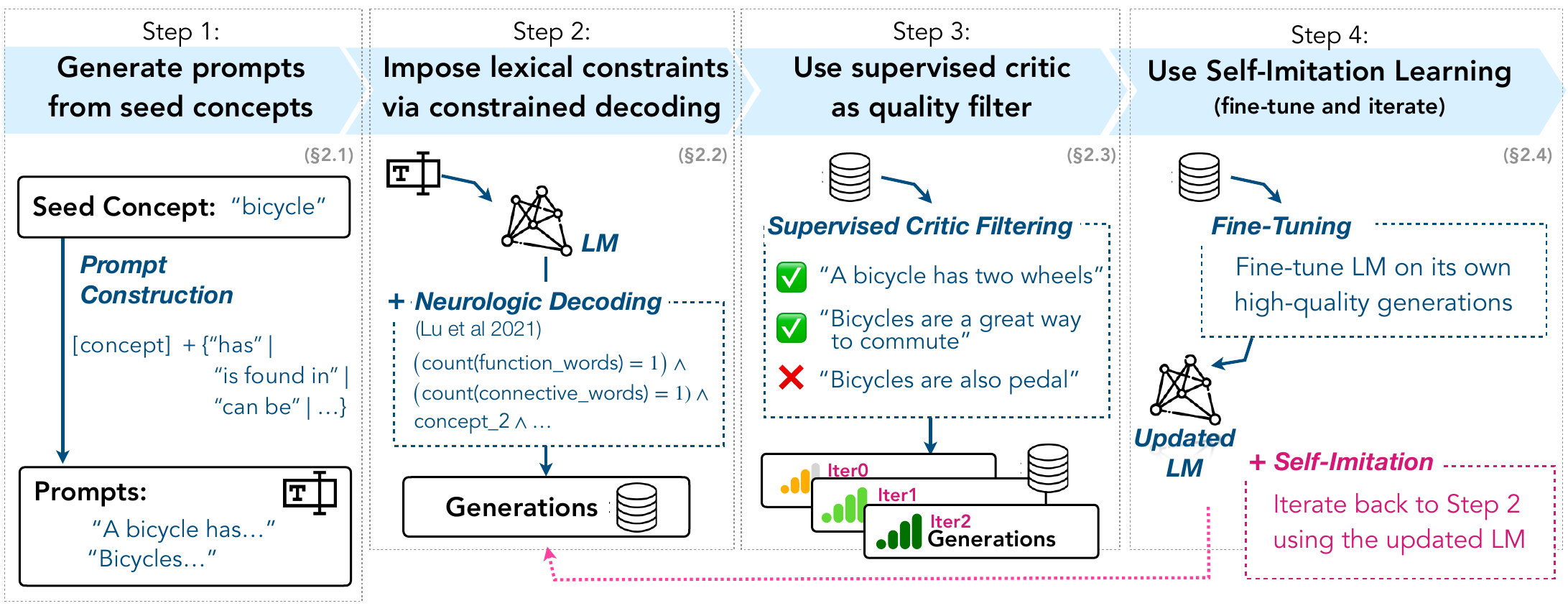}
  \caption{\framework is specifically designed to elicit \textit{generics}---general statements about the world. \framework works by collecting a list of concepts and generates generics using Neurologic Decoding to constrain generations at decoding time. To ensure quality, \framework includes the use of a supervised critic to filter out false generations. The quality of the generations is further improved via iterative self-imitation learning whereby the language model is finetuned on the high-quality generics selected by the critic.
  }
  \label{fig:overview}
\end{figure*}

We study the generation of commonsense knowledge from \gpttwo (a small LM) and compare that against \gptthree{}, a model that is orders of magnitude larger. Specifically, we focus on the task of generating \textit{generics} -- i.e. statements of commonsense knowledge about everyday concepts. While generics express general truths (e.g. ``birds can fly"), exceptions abound (e.g. penguins do not fly nor do sleeping or injured birds). Nonetheless, generics form the basis of how we express our commonsense about the world \citep{hampton2012generics, leslie2014carving}.

We present \framework, a new framework for generating generic statements from \gpttwo (depicted in Fig. \ref{fig:overview}).\footnote{I2D2: \textbf{I}terative \textbf{I}mitation and \textbf{D}ecoding for \textbf{D}istillation}
Out of the box, \gpttwo generations are anything but valid generics -- often being repetitive, trivial, or resembling narratives. 
The key breakthrough for overcoming this challenge comes from (i) \textbf{constrained decoding}: in which generations are controlled to satisfy manually constructed lexico-syntactic constraints using Neurologic Decoding \cite{lu2020neurologic}, and (ii) \textbf{self-imitation learning}: in which \gpttwo is iteratively fine-tuned on its own high-quality generations, automatically identified using a supervised critic model.

The marked disparity in scale makes the comparison between \framework and \gptthree seem like an impossible match. 
However, constrained decoding and self-imitation enable \framework to overcome this limitation and even surpass the quality of knowledge generated by \gptthree. 
We formulate a binary-classification task on a human-annotated test set of generic statements
and compare the precision-recall trade-off between \framework and Instruct-\gptthree by ranking statements using their critic and perplexity scores, respectively.\footnote{We use Instruct-GPT 3's \texttt{text-davinci-001} model in our experiments. In the rest of this paper, \gptthree refers to this model, unless stated otherwise.} 
\framework achieves an average precision of $0.92$ and outperforms Instruct-\gptthree, which has an average precision of $0.82$. 
Next, we show that iterative self-imitation learning dramatically improves the accuracy of generations from \gpttwoxl, even before applying the critic; increasing from 45\% $\xrightarrow[]{}$ 58\% $\xrightarrow[]{}$ 62\% over three iterations. 
Finally, we construct \corpus{} -- a knowledge resource of generic statements generated by applying \framework to 40K everyday concepts. Compared to GenericsKB \cite{bhakthavatsalam2020genericskb},  \corpus{} is judged by humans to be more accurate (75\% GenericsKB vs. 90\% \framework) while being larger (over 2X) in scale. Unlike GenericsKB, which was created through information extraction over text, \framework can provide commonsense knowledge for unseen concepts on-demand.

\section{The \framework{} Framework} 
\label{sec: methods}

\begin{table}[h]
    \centering
    \small
    \begin{tabular}{c}
        $[\text{Generally} | \text{Typically} | \text{Usually}]?\ [\text{a} | \text{an} | \text{the}]?$ \\ $\texttt{ <noun\_phrase> }\ \texttt{<relational phrase>}$
    \end{tabular}
    \caption{Template for automatically constructing morpho-syntactically varying prompts. `$?$' denotes the group of words is optional and `$|$' denotes the logical OR operator}
    \label{regex}
\end{table}

\framework{} is a new framework for automatically generating generic statements using pretrained language models. Our language model of choice is \gpttwoxl. 
However, any auto-regressive language model can be used within \framework.\footnote{In the rest of the paper, \framework{} refers to \framework{} using \gpttwoxl.} 

\framework{} generates generics in four stages. 
First, in \textbf{prompt construction}, we collect seed concepts (e.g. \textit{bicycle}) and automatically construct several morpho-syntactically varying prompts (e.g. \textit{``A bicycle has \ldots"}) (\S\ref{sec:prompts}) for each concept. The prompts are used as inputs to \framework.
Second, we employ \textbf{constrained generation} to control the style of text generated from the pre-trained LM at to mimic the style of generic statements(\S\ref{sec:constrained}).
Third, a supervised critic is used to \textbf{filter} out false and ill-formed generations (\S\ref{sec:critic}). 
Finally, the language model is finetuned on its own high-quality generations selected by the critic in an \textbf{iterative self-imitation learning} setup (\S\ref{sec:self-imitation}). 
Figure \ref{fig:overview} illustrates the overall framework.

\subsection{Prompt Construction}
\label{sec:prompts}

\paragraph{Source of seed concepts:}
Our first set of concepts for generating generic knowledge is common noun phrases (e.g. ``fruits"), selected from two resources: GenericsKB \citep{bhakthavatsalam2020genericskb} and ConceptNet \citep{speer2017conceptnet}. 
From GenericsKB, we retrieve all noun phrases for which there are at least five generic statements in the resource, resulting in a total of 8.5K seed concepts.\footnote{GenericsKB was found to consist of uncommon or specialized terminology (e.g. orpiment) that are not conducive for commonsense generation.  Therefore, we select nouns with at least five generic statements so that the collected nouns are those that are capable of forming commonsense generics} 
From ConceptNet, we retrieve noun phrases associated with the types \texttt{artefact} and \texttt{human}, identified based on hypernymy relationships to the corresponding WordNet senses. 
These lists are then manually vetted for validity to compile 
a shortlist 
totaling 
1.4K seed concepts.\footnote{We choose \texttt{human} and \texttt{artifact} as much commonsense knowledge centers around these types. 
The list of concepts can be extended to other types as well (e.g. animals, natural phenomena) in the future.} 

Our second set of seed concepts is high-level human goals (e.g. ``get better at chess") obtained from two sources: ProScript \citep{sakaguchi2021proscript} and ATOMIC \citep{sap2019atomic}. 
We extract all \texttt{goals} that appear in the ProScript training data. 
From ATOMIC, we extract all \texttt{base events} and filter out hypothetical ones (e.g. ``PersonX expects to win'') based on an exclusion list (Appendix \ref{appendix:selection_concept_and_goals}). 

To scale the number of seed concepts we prompt GPT-3 \cite{brown2020language} with a \textit{set-expansion template}, which is a prompt template for GPT-3 to generate items similar to a given set of items; see more details in Appendix \ref{appendix:gpt3Expand}. 
Overall, after GPT-3 based expansion, we have 39K seed concepts, consisting of 26K noun phrases and 13K goals. 
Note that GPT-3 is only used for seed expansion and not for the generics generation.

\paragraph{Morpho-Syntactically Varying Prompts:}
We programmatically construct a large number of morpho-syntactically divergent prompts for each concept to facilitate the generation of a diverse set of generics. %
Prompts for noun phrases 
are constructed based on the template shown in Table \ref{regex}.

\begin{figure*}[t!]
\centering
  \includegraphics[width=1\textwidth]{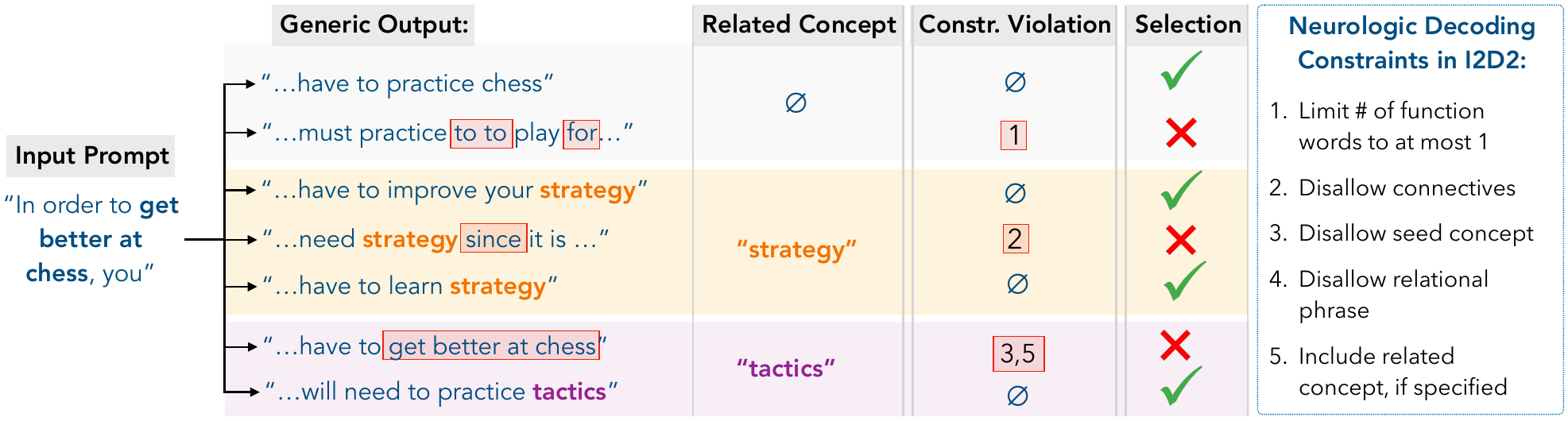}
  \caption{Example outputs of \framework for the concept ``get better at chess". We add constraints to our constrained generation algorithm to include the related concept.
  }
  \label{fig:constraints}
\end{figure*}

Each concept is paired with a \textit{relational phrase}, e.g. ``can be", ``is found in", from a manually constructed list; Appendix \ref{appendex:relations} presents more details.
Inspired by \citet{leslie2008generics}, we prefix adverbs (such as ``generally'', ``usually'', and ``typically'') to the prompts. We find, empirically, that these prefixes encourage the language model to generate general statements, instead of long-form, narrative-like text. 
An article is optionally prefixed before the concept for grammaticality. 
For a given (concept, relational phrase) pair, we construct all prompt combinations according to the template above and choose the one with the lowest PLM (\gpttwoxl in our experiments) perplexity. 
For the goal seed concepts, from each goal we create four separate prompts by prepending each of these prefixes: ``In order to", ``Before you", ``After you", and ``While you". 

\paragraph{Source of related concepts:} NLP applications often require knowledge that connects two concepts together in some given context. For example, to solve a QA problem, it might be important to have background knowledge about the relationship between a ``hotel" and a ``credit card", e.g. ``At a hotel, credit cards can be used to make a payment". We obtain concepts related to a seed concept from \gptthree using a custom template; see details in Appendix \ref{appendix:gpt3Related}. In Section \ref{sec:constrained}, we describe how \framework is able to generate such generic statements.

Finally, we filter out all prompts whose per-word perplexity under \gpttwoxl is above a threshold of 250. This allows us to apriori filter out ill-formed prompts such as ``Typically, a hall are planted at \ldots". This results in a total of 1.6M prompts.

\subsection{Constrained Generation using NeuroLogic Decoding} 
\label{sec:constrained}
\paragraph{Why Constrained Decoding:}
Small language models like \gpttwoxl  struggle with text degeneration \cite{holtzman2019curious}. Text generated can be trivial, repetitive, or long-winded %
resembling a narrative. In contrast, generic statements are simple and short \cite{Tessler2016APT}. The main challenge is to generate statements consistent with the linguistic style of generics, while using an inherently weak language model. To address this, we could either adapt the model to our task, through fine-tuning or apply novel decoding algorithms to substantially improve the generation quality. As the only resource of generic statements, GenericsKB \cite{bhakthavatsalam2020genericskb} 
could
be used for fine-tuning. But it primarily focuses on scientific concepts and, as we show in \S\ref{sec:experiments}, lacks diversity and scale. Crowdsourcing a new dataset from scratch is resource intensive. Thus, we focus on better decoding methods instead of relying on the standard top-p, top-k, or beam search algorithms.

\paragraph{What is NeuroLogic Decoding:}
NeuroLogic Decoding \cite{lu2020neurologic} enforces satisfaction of given constraints in generated text. It can handle any constraints---\textit{positive} (a given word must be included in the generation) or \textit{negative} (the given word must not be generated)---which can be expressed in conjunctive normal form. The constraint satisfaction problem is solved approximately using beam-search by introducing a high-penalty term for violating constraints.

\paragraph{NeuroLogic Decoding in \framework}
Our work is the first to use NeuroLogic Decoding for knowledge generation. 
The application of NeuroLogic to our problem is based on two key observations. First, we find that limiting the number of function words (e.g., ``in'', ``on'', ``of'') in a sentence implicitly controls its length. Next, excluding connective words (e.g., ``although'', ``since'', ``furthermore'') can make generations short and succinct. 

These logical constraints can be enforced at decoding time to steer the model toward desired text using NeuroLogic Decoding. 
We devise the following set of constraints, represented in CNF. 
Constraints are exemplified in Figure~\ref{fig:constraints} and further detailed in
\ref{appendix:constraints}. 

{\small
\vspace{-2em}
\begin{align}
    \big(& count(\texttt{function\_words}) \leq 1 \big) \nonumber\\ 
    &\land  \big( count(\texttt{connective\_words}) = 0)  \nonumber\\
    &\land \neg \texttt{source\_concept} \nonumber\\
    &\land \neg \texttt{relational\_phrase} \nonumber
\end{align}
}
Given the 1.6M programmatically constructed prompts and their associated constraints, we generate ten generations for each prompt using NeuroLogic Decoding applied to \gpttwoxl. Overall, we generate about 16M statements which must now be filtered to preserve quality.

\begin{figure*}[t!]
\centering
  \includegraphics[width=1\textwidth]{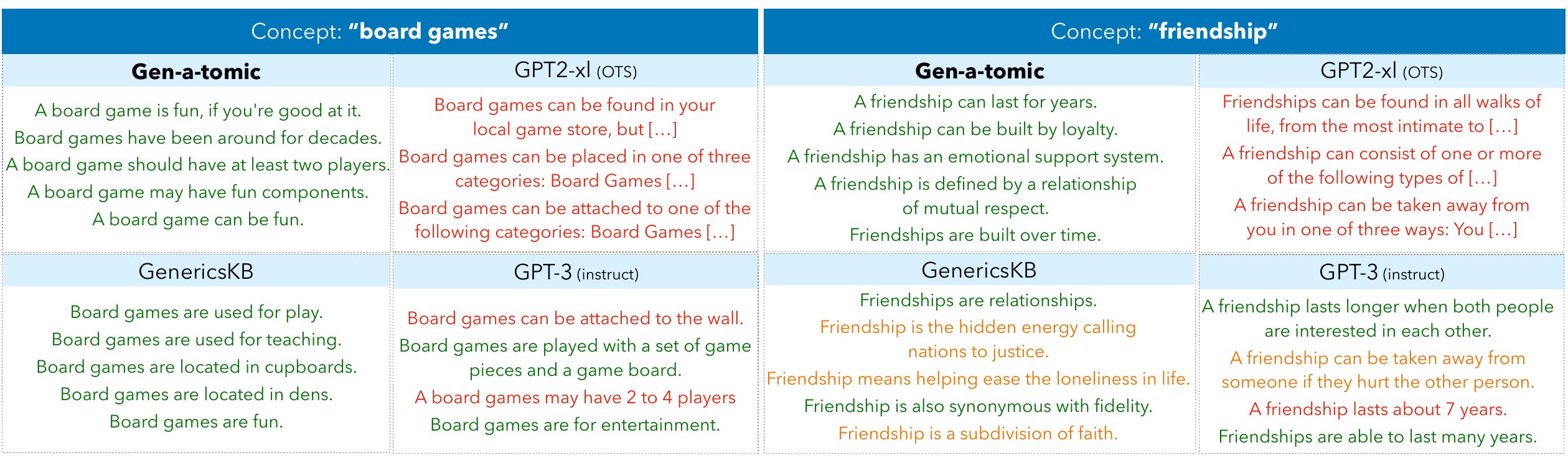}
  \caption{Examples of generics in \corpus and GenericsKB, and those generated by off-the-shelf GPT2-xl and GPT-3 instruct. Examples in green are good generics, red are bad generics, and orange are questionable ones.}
  \label{fig:examples}
\end{figure*}

\subsection{Supervised Critic}
\label{sec:critic}
LMs can generate hallucinations and false statements about the world \citep{ji2022survey}. We similarly observe invalid or false statements output by our constrained decoding method.
To address this, we train a supervised critic model to predict the veracity of a generation. 
We create a training set of $\sim$12K statements, with up to four sampled generations for each concept from a held-out set of $\sim$3K concepts. 
The labels for each generation are collected using the same procedure as the evaluation data, which is described in Section \ref{section:evaldata}. 
We train a RoBERTa-Large \citep{liu2019roberta} classifier as our critic model to identify valid generic statements.

\subsection{Self-Imitation Learning}
\label{sec:self-imitation}
\paragraph{Why Self-Imitation:}
NeuroLogic Decoding allows \framework to generate statements in the style of generics. But the deficiencies of using a weak language model are still apparent as the critic model has to discard a majority of the candidate statements due to their low quality. Intuitively, using a better language model should make it more likely for NeuroLogic to find higher-quality candidates. We posit that fine-tuning the language model on its own high-quality generations can make it better suited for 
knowledge generation by steering its distribution towards higher-quality samples.

\paragraph{What is Self-Imitation:} In the reinforcement learning literature, self-imitation learning \cite{oh2018selfimitation} is an actor-critic algorithm for learning to reproduce past \textit{good} actions of an agent in an environment. $\langle$\texttt{State}, \texttt{action}, \texttt{reward}$\rangle$ triples from past experience are stored in memory and an action taken in the past is chosen only when that action resulted in higher reward than expected. 

\paragraph{Self-Imitation in \framework:}
Our method closely follows self-imitation of \cite{oh2018selfimitation}, but uses a pre-trained language model as the \textit{`actor'} and a trained classifier as the \textit{`critic'} models. Moreover, we update the language model using the standard conditional language modeling objective, maximum likelihood. 
\framework is formally described in Algorithm \ref{alg:main}.

\renewcommand{\algorithmicrequire}{\textbf{Input:}}
\renewcommand{\algorithmicensure}{\textbf{Output:}}
\renewcommand{\algorithmiccomment}[1]{\hfill{\(\triangleright\)~#1}\par}

\begin{algorithm}[!t]
\small
\begin{algorithmic}[1]
\REQUIRE{$P_0$: a pre-trained language model\\ $\mathbf{X} = \{ (x_i, \mathcal{C}_i) \}$: Set of prompts and constraints\\ $\Omega(\cdot)$: a critic model\\ $N$: number of iterations\\ $\delta$: classification threshold}
\FOR{$k = 0, 1, \ldots, N-1$}
\STATE ${\mathcal{D}_k \leftarrow \{\}}$
\FOR{$(x_i, \mathcal{C}_i) \in \mathbf{X}$} %
\STATE ${\hat{\textbf{y}} = \arg \max_{\textbf{y} \in \cal{Y}} P_k (\textbf{y} | \textbf{x}) - \lambda \sum\limits_{c \in \mathcal{C}_i} \big(1 - c(\textbf{y})\big)}$  \algorithmiccomment{Constrained Decoding}
\STATE $\mathcal{D}_k \leftarrow \mathcal{D}_k \cup \{\hat{\textbf{y}}\}$ \algorithmiccomment{Add generations to data pool}
\ENDFOR
\STATE Train $\Omega$ on annotated samples of $\mathcal{D}_k$ \algorithmiccomment{Train Critic}
\STATE ${\tilde{\mathcal{D}}_k \leftarrow \{\textbf{y} | \textbf{y} \in \mathcal{D}_k \text{ and } \Omega(\textbf{y}) > \delta}\}$ \algorithmiccomment{Apply Critic}
\STATE ${ P_{k+1} \leftarrow \arg\max_{\theta} \E{} \log P_k(y_n | y_1,\ldots y_{n-1}); \textbf{y} \sim \tilde{\mathcal{D}}_k }$ \algorithmiccomment{Train LM on high-quality generations}
\ENDFOR
\end{algorithmic}
\caption{The \framework framework}
\label{alg:main}
\end{algorithm}

\section{Experiments and Results}
We describe results from our experiments comparing \framework with \gptthree, \gpttwoxl and GenericsKB in more detail below. Figure \ref{fig:examples} shows outputs sampled from these sources.
\label{sec:experiments}

\begin{figure}
\centering
\includegraphics[width=0.45\textwidth]{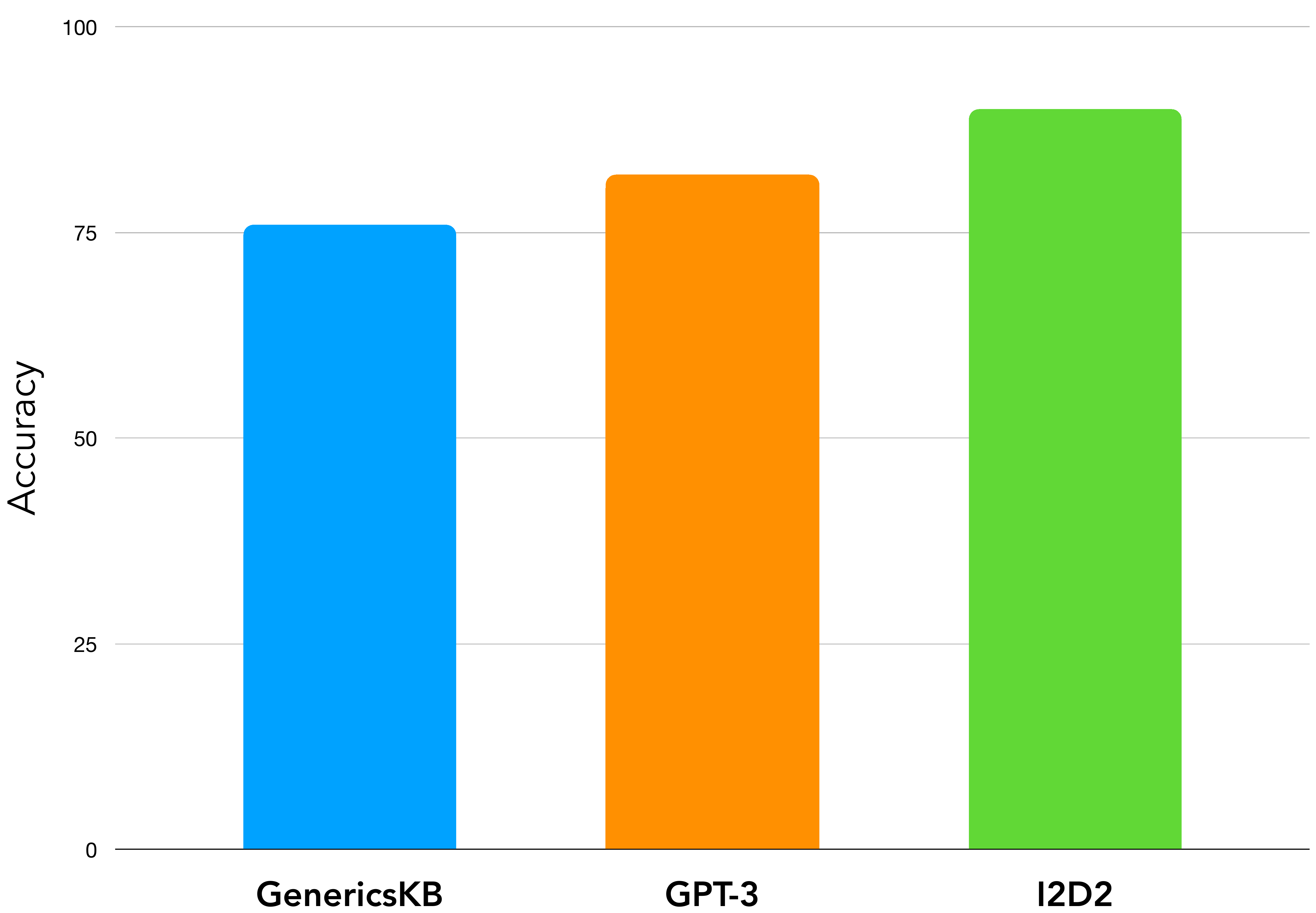}
\caption{The accuracy of \framework generations is higher than GPT-3 (a $100\times$ larger model) and GenericsKB}
\label{fig:res1}
\end{figure}   

\begin{figure*}[ht!]
    \centering
    \begin{minipage}[t]{.49\textwidth}
        \centering
        \includegraphics[width=0.8\textwidth]{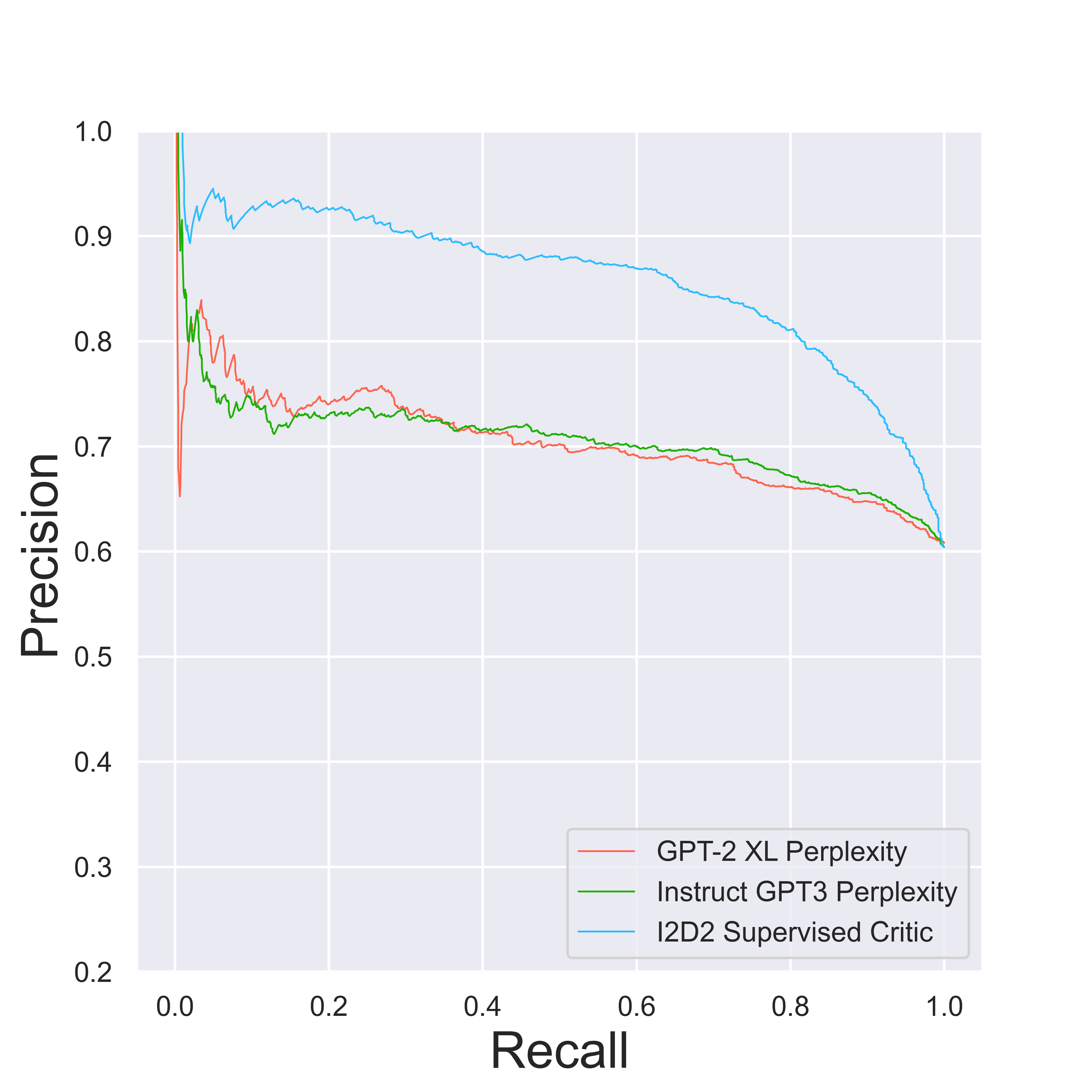}
        \caption{Comparing the PR Curve of \framework{} Critic and the language model perplexities based on GPT2-XL and GPT-3.} %
        \label{fig:prcurvea}
    \end{minipage}%
    \hfill
    \begin{minipage}[t]{0.49\textwidth}
        \centering
        \includegraphics[width=0.8\textwidth]{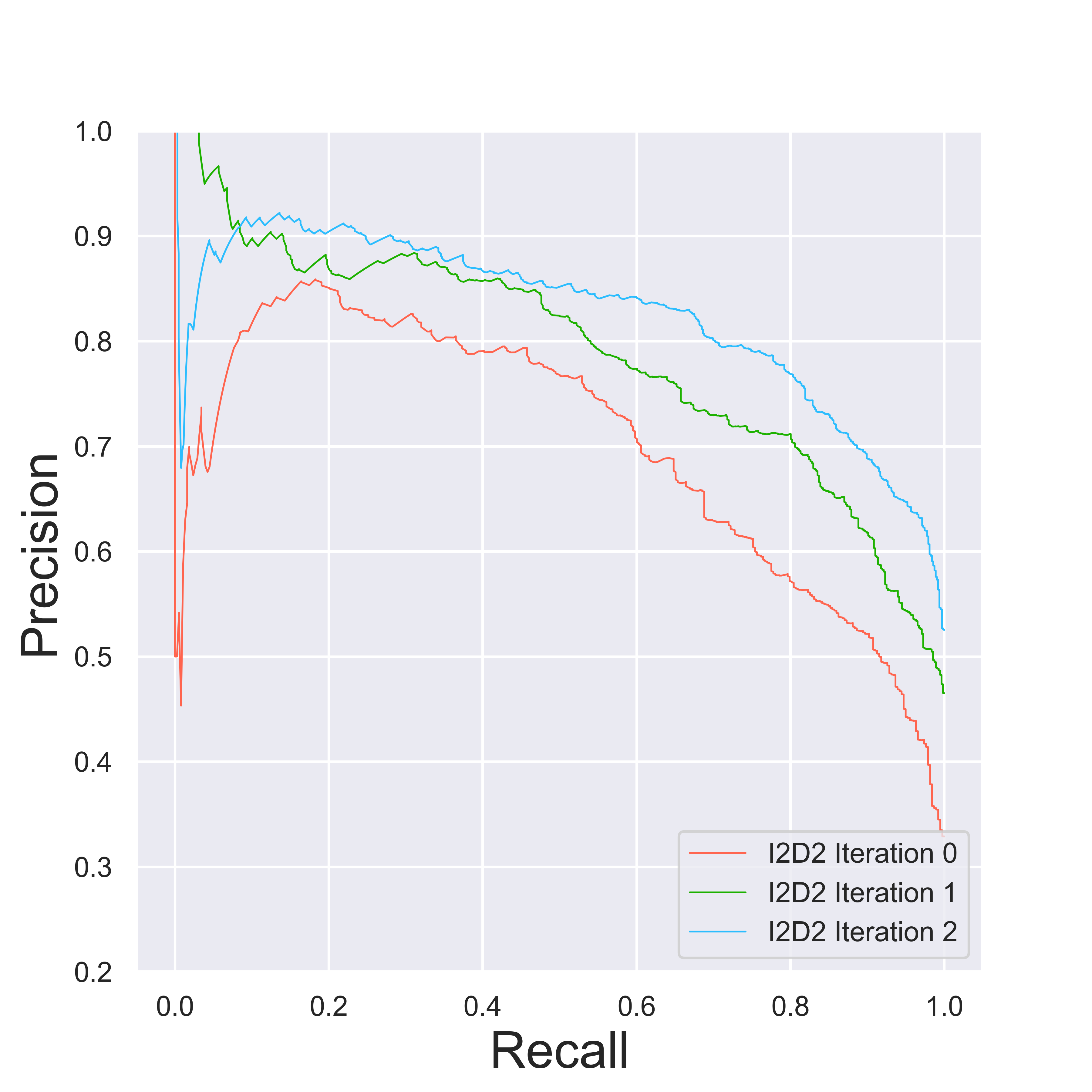}
        \caption{Comparing the PR Curve of \framework iterations 0 through 2}
        \label{fig:prcurveb}
    \end{minipage}
\end{figure*}

\subsection{\framework's generations are more accurate than GPT-3 and GenericsKB}
We compare the accuracy of generations from \framework, \gptthree, and GenericsKB (see Figure \ref{fig:res1}). The best accuracy achieved by \gptthree in our experiments is \textbf{82\%}. 
GenericsKB \citep{bhakthavatsalam2020genericskb} is a static resource of generic knowledge created through information extraction over three large text corpora: the Waterloo corpus, SimpleWikipedia, and the ARC corpus. This work released a large-scale dataset of 14M generations and a high-quality subset of 1M generic statements. We compare GenericsKB's best 1M against our corpus. We randomly sample 1K generic statements from GenericsKB and \framework and ask annotators on Amazon Mechanical Turk (MTurk) to rate the validity of the generic statement. We find that while only \textbf{76\%} of statements in GenericsKB were annotated as accurate, over \textbf{90\%} of statements in \framework were judged as valid.  
The results show that \framework is more accurate than GenericsKB, while being larger. \framework is also more accurate than \gptthree, while using $100\times$ fewer parameters in its model.

\subsection{\framework results\,in\,better\,generics\,than \gptthree}

\paragraph{Systems} We wish to compare how \gptthree, given the same set of prompts as our approach, can generate and identify valid generics. For a given prompt, we generate ten generations from each system. \gptthree is prompted in a few-shot manner with an instruction and six examples. We use different sets of few-shot examples for noun phrases and goals. Appendix \ref{appendix:gpt3Generations} further details the instruction and in-context examples provided to \gptthree. \framework, using a supervised critic, assigns a score to each generated statement. For \gptthree, we use the perplexity assigned to a generation as an indicator of validity. As an additional baseline, we also compute perplexity under off-the-shelf \gpttwoxl.

\paragraph{Evaluation Data}
\label{section:evaldata}
We set aside 300 concepts for evaluation. Each concept is associated with several prompts (on average 40). We generate ten generic statements for each prompt from \framework and \gptthree. Next, from all generations for a concept, we randomly sample four statements generated by each system. %
A generic statement is considered valid if it is a generally true statement about the world. Three annotators on MTurk rate the validity of each generated statement.\footnote{Annotators select one of four choices: \{true, false, don't know, garbled output\}.} Annotation template and instructions are detailed in Appendix \ref{appendix:annotationTemplate}. At least two out of three annotators agreed on a label 92.5\% of the time over all 4 statements.\footnote{We provide pairwise annotation agreement. Since our generations should ideally be valid, we produce a skew towards a single label, problematic for $\kappa$ \cite{feinstein1990high}.} %

\paragraph{Metrics} Given the human-annotated test set of generics, we compare the precision-recall trade-off between \framework and \gptthree. Each system assigns a score to each generic statement, allowing us to rank the statements from most to least likely to be a generic. Combined with the human annotations of the validity of a statement, we plot a precision-recall (PR) curve. It allows us to evaluate the accuracy of each system as the number of statements it outputs varies, which is important since different tradeoffs between quantity and quality of output may be desired for different application settings.

\paragraph{Results} Figure \ref{fig:prcurvea} shows the impact of including a supervised critic to identify valid generic statements. We find that \gptthree, while impressive, lags significantly behind our supervised critic in identifying which generic statements are valid. %
The off-the-shelf \gpttwoxl model is the worst at identifying valid generic statements. Perplexity alone is not a good indicator of what a valid generic is.

\framework uses both a generator and a discriminator. To evaluate the generator, we sample from its generations over the test set of prompts. For a given set of generations, human annotators judge whether the statement is true or false. We compute accuracy against human labels and use that as a metric to measure the quality of the generator.

\paragraph{The cautions against \gptthree comparison}
There are growing concerns in the research community about the lack of open availability of \gptthree. Several versions of \gptthree are available through an API, but the details of the training data used for each version are largely unavailable or underspecified. Direct comparison with \gptthree is, therefore, becoming increasingly challenging. In this work, we compare against the `text-davinci-001' version of the \gptthree model and note that newer models might do better. However, extracting the best performance from \gptthree is beside the point of our work. We believe that as a community, we must investigate alternative approaches that do not just rely on scale. Case in point, our results in \S\ref{ref:gpt3familycomp} demonstrate that the smaller \texttt{curie} version of \gptthree outperforms the much larger \texttt{davinci} version, through better training.

\subsection{\framework gets better through iterative self-imitation learning}

\paragraph{Systems} For self-imitation learning, we generate a large corpus of generations and filter out invalid statements using the supervised critic to yield a ``purified" subset. We compare generations from \framework using off-the-shelf \gpttwoxl and outputs from two additional iterations of fine-tuning. 

\paragraph{Evaluation Data} We use the same held-out test set of prompts for this experiment. 

\paragraph{Metrics} Here, we evaluate the accuracy of the generations before applying the supervised critic.

\paragraph{Results} We show that a language model gets iteratively better as it gets finetuned on its own high-quality generations over each iteration. The raw accuracy of the generations, before applying the critic, improves from 45\% $\rightarrow$ 58 \% $\rightarrow$ 62\% over three iterations. We also compare the precision-recall trade-off between the three iterations. Figure~\ref{fig:prcurveb} shows the effectiveness of self-imitation learning over three iterations.

\subsection{\corpus is more diverse than GenericsKB}
\corpus{} is a large set of generic statements, but some of these may be semantically equivalent to one another.  Since exact quantification of semantically distinct statements in the dataset is intractable, we employ a survey method called Mark and Recapture (MnR) \cite{seber1982estimation,usgs2018capture} commonly used by ecologists to estimate a large population size via sampling. 
This method captures individuals of a population in two (or more) stages. In the first capture, the 
generics capture (i.e., sampled) are \textit{marked} and released. At a later capture, the number of \textit{recaptured} generics\footnote{A recapture is determined by the second sample's BLEU score with respect to the already captured.} are counted and the population size estimated. Then, we employ the Chapman estimator for MnR \cite{brittain2009estimators,chapman1951some} to estimate the population size of unique generics in the dataset. More details can be found in Appendix~\ref{appendix:mnr}. 

We compare the estimated \textit{per concept} average count of unique generics for GenericsKB and \corpus. Overall, we find that \corpus includes at least triple the amount of generics per concept compared to GenericsKB. We also observe that the estimated unique generics per concept is higher for the best cuts of the \corpus dataset.  Experiments with embedding-based similarity methods yielded similar results.

\begin{figure}
\centering
\includegraphics[width=0.48\textwidth]{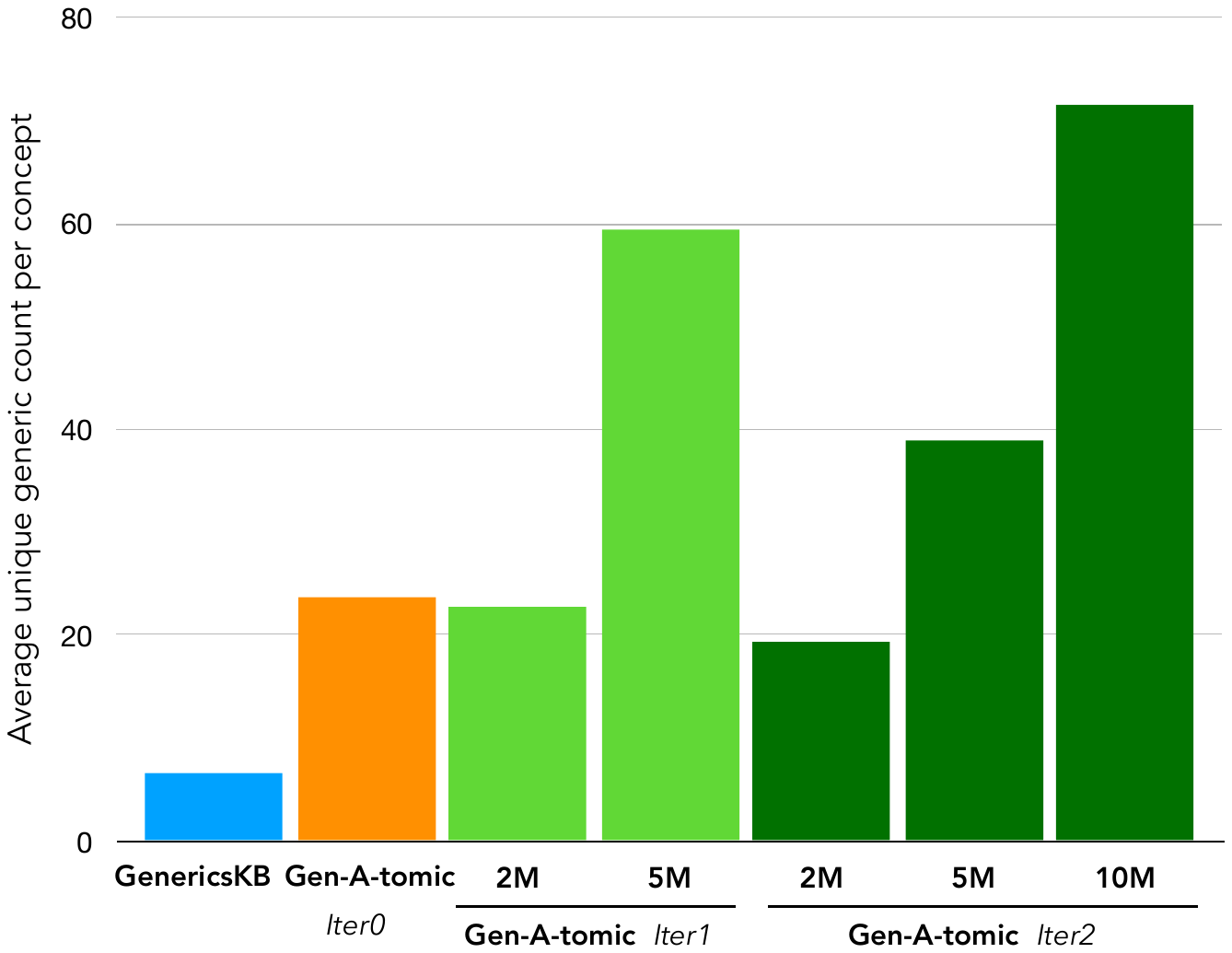}
\caption{Compared to GenericsKB, the estimated average unique number of generics per concept is higher for any version of \corpus.}
\end{figure}

\subsection{Smaller, better-trained versions of GPT-3 outperform larger ones}
\label{ref:gpt3familycomp}
We compare three versions of the GPT-3 model available on the OpenAI API: \texttt{\small davinci}, \texttt{\small curie-instruct} and \texttt{\small davinci-instruct} \cite{ouyang2022training,brown2020language}. Interestingly, we find that the curie-instruct model, despite being a much smaller model, generates more valid generic statements compared to the much larger davinci model. The instruct models (including curie-instruct) were trained using reinforcement learning on human feedback. The accuracy (validity) of statements generated by the three GPT-3 models on the same set of test prompts are 53.3\% (davinci), 60.6\% (curie-instruct), and 81.9\% (davinci-instruct). These results further demonstrate that better training can result in smaller models performing better than larger models.

Our work adds to the growing body of evidence from recent work that large language models have not been trained optimally \cite{kaplan2020scaling} and it would be worthwhile to look for better training strategies to achieve high performance using smaller, affordable, greener models.

\section{Related Work}
\paragraph{Generics}
Generics like ``dogs are friendly'' describe observed ``truths'' or defaults about the world for which exceptions can be found (e.g., not all dogs are friendly in practice).
Generics have been studied quite extensively in philosophy, linguistics, and psychology. While they are clearly important to human reasoning, in particular, to non-monotonic reasoning \cite{carlson1995generic,Pelletier1997GenericsAD}, they have also been long debated for their puzzling properties which renders them difficult to formally analyze \cite{Leslie2012-LESG-3,leslie2008generics, hampton2012generics, liebesman2011simple}. %
\citet{bhakthavatsalam2020genericskb} demonstrated the usefulness of generics in language understanding by providing generic statements to text models and showing improvement on question-answering and explanation generation. However, being a static resource, GenericsKB cannot provide knowledge for unseen concepts. To be useful across a wide range of tasks and datasets, a more comprehensive resource of generics is required. \framework{} can generate generics for arbitrary concepts and even generics relating two concepts---a feature unique to \framework{}. \framework{}  can is easily extensible temporal (``during a cold night, people need a blanket") or comparative (``a tennis ball is smaller than an office chair") generic knowledge, leading to a more comprehensive commonsense knowledge model. %

\paragraph{Commonsense Knowledge}
Various methods for representing commonsense knowledge have been proposed in the literature. 
ConceptNet \cite{speer2017conceptnet} focused on the conceptual commonsense relationship among various concepts and entities in their knowledge graph. Atomic \cite{sap2019atomic} and Atomic2020 \cite{hwang2021comet} have offered symbolic commonsense knowledge graphs representing relational inference focusing on the ``If-Then'' (cause-effect) reasoning.
Fine-tuned on Atomic, Comet \cite{bosselut2019comet} has offered a neural knowledge model that can reason about situations beyond the symbolic knowledge graphs. Unlike our current framework, however, previous commonsense knowledge models typically only handled data in the form of structured triples and were predominantly focused on commonsense about events. \framework{} is the first knowledge model focused on generic knowledge expressed in natural language.
Uniquely, we also provide a critic model that can filter invalid or ill-formed generations.

\paragraph{Symbolic Knowledge Distillation}
Collecting high-quality knowledge at scale has been a longstanding challenge. The traditional way is to collect by human annotation \cite{speer2017conceptnet,sap2019atomic}%
, which can be time-consuming and 
expensive. \citet{bhakthavatsalam2020genericskb} extracted generics by filtering and cleaning based on 1.7B sentences from three large text corpora. However, manually constructed resources and resources extracted from large corpora 
can be difficult to extend. Recent works showed that pre-trained language models can be a good source of knowledge \cite{west-etal-2022-symbolic,zhang2022opt}. Symbolic knowledge distillation (SKD) \cite{west-etal-2022-symbolic}, for instance, has generated even-centric inferential knowledge from \gptthree and distills it into \gpttwo. While these methods present promising results, they primarily rely on using \gptthree and only handle knowledge about events in a structured triple format. \framework{}, on the other hand, relies only on \gpttwo's own generations to improve itself and generates knowledge in natural language.

\paragraph{Self-Imitation Learning}
Self-imitation learning \cite{oh2018selfimitation} was proposed as a reinforcement learning method in which an agent learns to replicate past good actions. 
More recently, a similar approach was applied in dialog models \cite{lamda,xu2022learning} and code generation \cite{haluptzok2022language}. However, recent applications have 
relied on
models 
much larger 
than \gpttwoxl used in \framework{}. Moreover, while \cite{haluptzok2022language} have explored the idea of self-imitation learning in language models, their method relies on a compiler that is, by definition, 100\% accurate. Instead, 
the supervised critic in \framework{} can be noisy, especially for identifying generics, which have paradoxical properties that make its formalization very difficult \cite{mari2012genericity}. 
We also show that self-imitation learning is beneficial when done over multiple iterations. In principle, \framework{} could be improved iteratively through a life-long learning process. But, under what conditions would the performance gains plateau is an interesting open future research question.

\section{Conclusion}
\label{sec:conclusion}

We present \framework --- a novel framework for generating generic knowledge from language models using constrained decoding and self-imitation learning. \framework, while using orders of magnitude fewer parameters, can still outperform \gptthree at the task of generating high-quality generic statements. We also show that \corpus is higher-quality, larger-scale, and more diverse than the static GenericsKB dataset. \framework provides on-demand access to generic knowledge that can bridge the gap in commonsense knowledge, often observed in even the largest LMs available today.

\section{Acknowledgements}
We thank our colleagues on the Beaker Team at the Allen Institute for AI for helping with the compute infrastructure. This work was supported in-part by DARPA MCS program through NIWC Pacific (N66001-19-2-4031). We thank the reviewers and ACL area chairs for their valuable feedback that made our work better. 

\section*{Limitations}
\paragraph{Comparison with \gptthree:} There are growing concerns in the research community about the lack of open availability of \gptthree. There are several versions of the model and the details of the training data used for each version are largely unavailable. Direct comparison with \gptthree is, therefore, becoming increasingly challenging. In this work, we compare against the `text-davinci-001' version of the \gptthree model and note that newer models might do better. However, extracting the best performance from \gptthree is beside the point of our work. We believe that as a community, we must investigate alternative approaches that do not only rely on scale. 

\paragraph{Undesirable Generations:} Language models, large and small, have been shown to be prone to generating toxic text \cite{realtoxicity}. \framework relies on \gpttwoxl could also potentially generate toxic statements. While the trained critic model is able to filter out most toxic generations, we estimate the proportion of undesirable generations using the Delphi \cite{jiang2021delphi} model. We find that $\sim 1.3\%$ of the generations may not be morally acceptable, either because the statements are not accurate, not verifiable, too restrictive, or they are potentially toxic.

\paragraph{Self-Imitation Iterations: } In this work, we only try two iterations of self-imitation due to resource constraints. Exploring the effects of more self-imitation iterations is left for future work. But, based on the performance improvements we observed after two iterations, we hypothesize that the improvements could diminish with each future iteration.

\paragraph{Runtime Efficiency} A batch of 32 generations from \framework takes ~3mins on a single RTX A6000 GPU. NeuroLogic Decoding is the most computationally expensive component. As constrained decoding methods become more efficient, the runtime of I2D2 will also improve. Our focus in this work is to study the quality of generations and we leave runtime efficiency improvements to future work.

\section*{Ethical Statement}

\paragraph{Crowdsourcing:} Annotations were conducted on Amazon Mechanical Turk. For this project, we obtained an exemption through our institution's internal IRB. We do not retain nor publish deanonymizing information such as MTurk IDs. Throughout the project, we maintain an average hourly rate of \$15/hour for all our evaluations. More detail on annotation is available in Appendix~\ref{appendix:annotationTemplate}.

\paragraph{Intended Use:}  The framework \framework is intended to enable further research in knowledge generation using a smaller and openly available language model like \gpttwo. As discussed towards the end in \S\ref{sec:experiments}, large language models like \gptthree are indeed more capable of generating commonsense knowledge than off-the-shelf \gpttwo, but they as unavailable for open use. This work seeks to expedite a more sustainable yet high-quality generation using smaller models that are accessible to all.

\corpus can be used as a resource of static knowledge for downstream applications in NLP. As discussed in the Limitations section above, there may exist a small number of generations that may be considered toxic and harmful for use. Therefore, we emphasize that the dataset should be used for research purposes only. Moreover, because the dataset has been vetted by crowdworkers originating from North America, the knowledge of the retained generics in \corpus is most strongly representative of generalizations or `truths' of the English-speaking Western, specifically North American cultures. Extending it to encompass a more diverse set of world knowledge is a topic of our future research.

\bibliography{anthology,custom,__refs}
\bibliographystyle{acl_natbib}

\clearpage
\appendix
\section{Appendix}

\subsection{Selection of Concepts and Goals}
\label{appendix:selection_concept_and_goals}

\paragraph{ConceptNet concept selection} To select ConceptNet concepts, we first build a list of \texttt{artefact} and \texttt{human} terms from WordNet by hierarchically traversing the hypernymy hierarchy (by depth) starting from the \textit{artefact\%1:03:00} and \textit{person\%1:03:00}, respectively. We then select ConceptNet concepts that belong to the lists in the same order as the WordNet list. The concepts in the list are sequentially evaluated manually to build a total of 1K \texttt{artefact} and 400k \texttt{human} terms. The totaling 1.4k concepts are then used as seed ConceptNet concepts.

\paragraph{ATOMIC goal selection} To select goals from ATOMIC, we obtain the complete list of \texttt{base events} (e.g. ``PersonX adopts a cat''). We drop the ``PersonX'' prefixation and all mentions of ``Person" (e.g. ``PersonY'', ``PersonZ''). Additionally, because we want to select for goals that are achievable, we remove all irrealis or hypothetical situations (the described situation or event has not taken place). More specifically, we filter out all events with the verbs `need', `want' `wish', `hope', `dream', `expect', `imagine', `mean', and `plan'; negated events (e.g. ``PersonX does not get the job''); and events modified by modals that indicate permission or obligation (e.g. `should'). In this manner we arrive at a list of 8.5K goals from ATOMIC.

\subsubsection{GPT3 for Set Expansion}
\label{appendix:gpt3Expand}
We develop a template for set expansion based on GPT3. 
\begin{quote}
\small
Generate more concepts.\\
1: $<$sampled concept 1$>$\\
2: $<$sampled concept 2$>$\\
3: $<$sampled concept 3$>$\\
4: $<$sampled concept 4$>$\\
5: $<$sampled concept 5$>$\\
6:
\end{quote}
Set expansion is done in several iterations. 
We define $K$ as the number of new concepts to be found in each iteration. We construct a prompt as shown above by sampling five concepts. We get five outputs for each prompt. We skip concepts whose generation perplexity is lower than a set threshold (8 in our experiments). Thus, at most five new concepts are found with each call to the OpenAI API. At the end of each iteration, newly found concepts are added to the seed list of concepts. This iterative process allows us to slowly expand the original list with new related concepts.

\subsubsection{List of relational templates}
\label{appendex:relations}
Noun phrases are combined with one of the following verb phrases if they are obtained from GenericsKB:
\begin{quote}
\small
are\\
is\\
have\\
can\\
has\\
should\\
produces\\
may have\\
may be\\
\end{quote}

If a noun phrase is obtained from ConceptNet, we expand the templates available in the (file ``templates.txt" attached in supplementary material.

\subsubsection{Template for obtaining related concept}
\label{appendix:gpt3Related}

Related concepts for a given concept are also obtained from GPT3. We use the following prompt:
\begin{quote}
\small
    Generate five words that are related to the given word.\\ \\
    Word: hotel\\
    Related Words:\\
    1: Credit card\\
    2: Fee\\
    3: Resort\\
    4: Parking lot\\
    5: Reception\\
    ==\\
    Word: $<$given concept$>$\\
    Related Words:
\end{quote}
\gptthree generates five related concepts for each given word. 

\subsubsection{Constraints for Neurologic Decoding}
We use four sets of constraints for Neurologic Decoding:
\begin{align}
    \big(& count(\texttt{function\_words}) \leq 1 \big) \nonumber\\ 
    &\land  \big( count(\texttt{connective\_words}) = 0)  \nonumber\\
    &\land \neg \texttt{source\_concept} \nonumber\\
    &\land \neg \texttt{relational\_phrase} \nonumber
\end{align}

\texttt{function\_words} comprises of \texttt{\{``in",
    ``on",
    ``of",
    ``for",
    ``of",
    ``at",
    ``in",
    ``anybody",
    ``it",
    ``one",
    ``the",
    ``a",
    ``that",
    ``or",
    ``got",
    ``do"\}}.

\texttt{connective\_words} comprises of \texttt{\{``without",
    ``between",
    ``he",
    ``they",
    ``she",
    ``my",
    ``more",
    ``much",
    ``either",
    ``neither",
    ``and",
    ``when",
    ``while",
    ``although",
    ``am",
    ``no",
    ``nor",
    ``not",
    ``as",
    ``because",
    ``since",
    ``although",
    ``finally",
    ``however",
    ``therefore",
    ``because",
    ``consequently",
    ``furthermore",
    ``nonetheless",
    ``moreover",
    ``alternatively",
    ``henceforward",
    ``nevertheless",
    ``whereas",
    ``meanwhile",
    ``this",
    ``there",
    ``here",
    ``same",
    ``few",
    ``1",
    ``2",
    ``3",
    ``4",
    ``5",
    ``6",
    ``7",
    ``8",
    ``9",
    ``0",
    ``similar",
    ``the following",
    ``by now",
    ``into"\}}

We additionally add the \texttt{source\_concept} and the associated \texttt{relational\_phrase} that were used to compose the prompt.

\label{appendix:constraints}

\subsubsection{Human Evaluation}
\label{appendix:annotationTemplate}

\begin{figure*}
    \centering
    \includegraphics[width=0.8\textwidth]{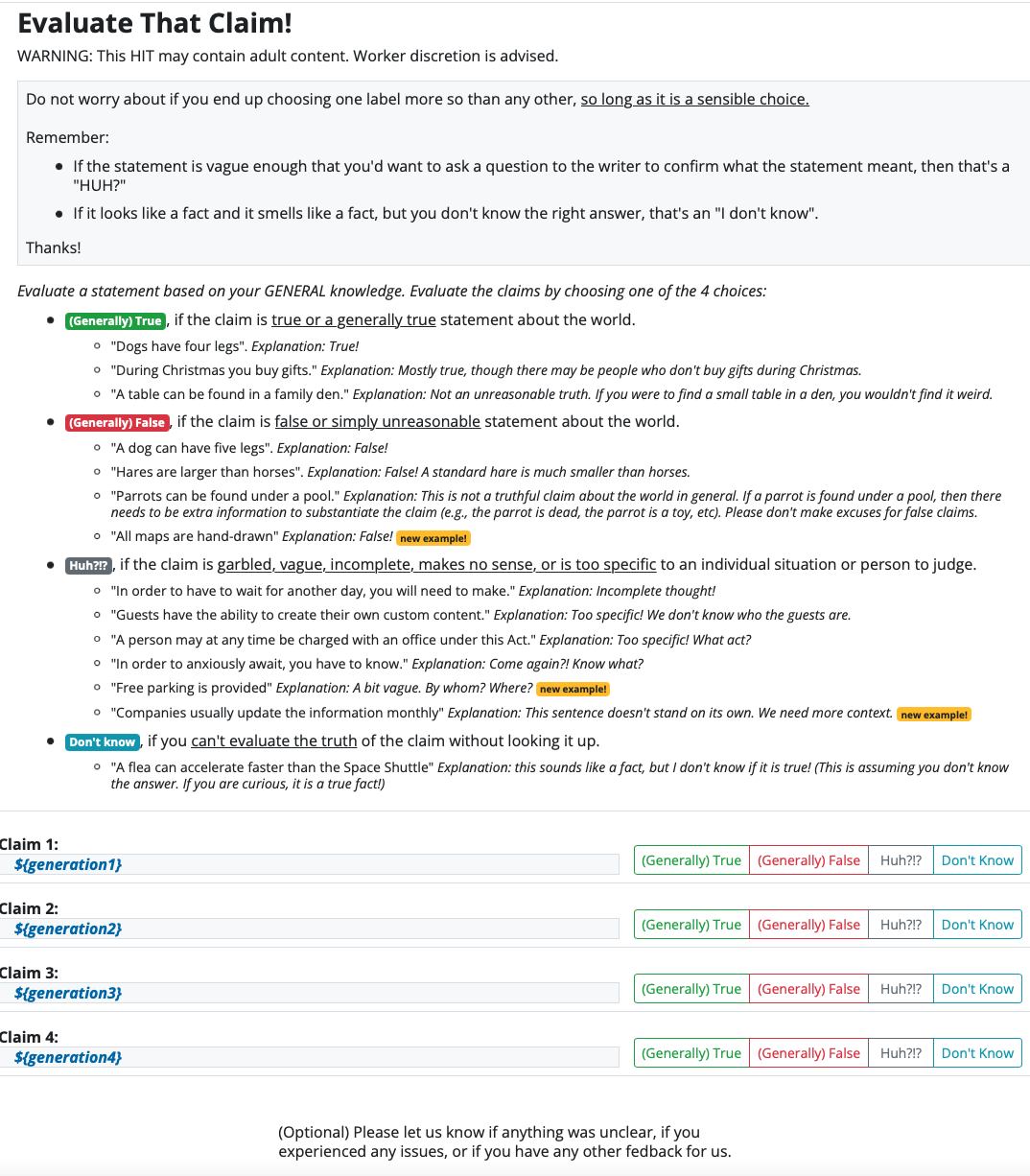}
    \caption{A screenshot of the template used for obtaining annotations on the Amazon Mechanical Turk platform.}
    \label{fig:turkTemplate}
\end{figure*}

All human evaluations were conducted through the Amazon Mechanical Turk (IDs) platform. We sourced our annotators from a pool of 168 crowdworkers manually selected from best performing workers in a round of open paid qualification. For the evaluation, the workers asked to rank model predictions on a 4-point validity likert scale. The Figure \ref{fig:turkTemplate} shows a screenshot of the annotation template with the full set of instructions used for collecting the training set and for evaluation of generic statements. Throughout the entirety project, we maintained an average of \$15/hour pay rate. 

We obtained IRB exemption for our evaluation from our institution's internal institutional review and ethics board. We did not collect any deanonymizing information nor do we publish with our dataset sensitive information such as MTurk IDs in full compliance to the exemption clauses found in 45 CFR 46.104(d)(2,3). Additionally, the extent of the crowdsourcing for the present work is limited to judgments based on world knowledge, we have no reason to believe that our crowdsourcing set up posed harm or discomfort beyond the minimal risk as defined by 45 CFR 46.102(i). Our exempted status does not require for us to use consent forms with our crowdsourcing.

As shown in the screenshot, the evaluations were conducted in English. Although we did not collect demographic information from the crowdworkers, our previous internal study from 2020 tells us that over 90\% of our annotators are English speakers from the US. Thus, the evaluation received as to the validity of the generic statements most strongly reflect the North American

\subsubsection{\gptthree generics generation template}
\label{appendix:gpt3Generations}

Generics are generated from \gptthree using the following template:

\begin{quote}
    Generate statements that are generally true in the real world.\\
    An apple is a fruit.\\
    Violins are used for music.\\
    Aardvarks are mammals.\\
    Accidents cause injuries.\\
    Protein is made of amino acids.\\
    Apples can be red or green.\\
    $<$ test prompt $>$
\end{quote}
We generate ten continuations for the prompt above.

\subsubsection{Mark-and-Recapture Details}
\label{appendix:mnr}
We use MnR to estimate the \textit{unique population size} of our large datasets thereby gauging the diversity of the dataset. For our implementation of MnR, we perform two random captures using a sample size of 30\% (of the total dataset size) at each capture. A generic in the second capture is considered a \textit{recapture} (i.e., individual seen in the first capture) if it exceeds a textual similarity threshold (BLEU score $>0.85$) with the \textit{generics of the same concept} from the first capture as the reference. The threshold was determined via several rounds of experimentation and manual evaluation to determine a reasonable level of textual similarity.
Then, we employ the Chapman estimator for MnR \cite{brittain2009estimators,chapman1951some} to estimate the population size of unique generics in the dataset. 

\subsubsection{Categories of generated Generics}
In our preliminary experiments, we collected crowdsourced annotations to label generated generics with categories derived primarily from \cite{leslie2008generics}. We found that the task was extremely challenging for non-expert crowdworkers. For example, recognizing “mosquitoes carry the West Nile virus” as a \textit{striking} generic requires a domain knowledge that often falls outside common knowledge.  As a result, we encountered low inter-annotator agreement scores leading us to not include them in the main discussion. 
However, based on samples from the first iteration of I2D2, we observed the following distribution of categories of generics: 
\begin{enumerate}
    \item semi-definitional (e.g., “laser produces a beam of light”): 45%
    \item characterizing: 35%
    \item striking or majority: 20%
\end{enumerate}

\subsection{Regarding License for \framework and \corpus}
The codebase for \framework will be licensed and released under the Apache License 2.0. The \corpus will be licensed under CC-BY.

\subsection{Responsible AI Checklist}
\paragraph{Number of parameters used} \framework mainly uses two models: \gpttwoxl with 1.5B parameters and RoBERTa-large with 354M parameters.

\paragraph{Total Computation Cost} 
We use Nvidia A6000 GPUs (with 48G RAM) in our experiments. The bulk of the computation cost is in executing constrained decoding over a large number of prompts to create \corpus. We can generate 10 generations each for 32 prompts in 2 mins. Overall, to generate 16M generic statements, we need about $\sim$1500 GPU hours. That said, creation of the large corpus is a one-time cost. \framework is readily applicable as a knowledge model that can be queried on-the-fly. Retraining the language model takes $\sim$24 GPU hours.

\paragraph{Hyperparameters} We use the following hyperparameters for different components:

\textbf{For constrained decoding: } 
\begin{quote}
 \texttt{batch size} 32, \\
\texttt{beam size} 10, \\
\texttt{max generation length} 30, \\
\texttt{min generation length} 2, \\
\texttt{length penalty} 0.1    \\
\end{quote}

\textbf{For training the critic model: } 
\begin{quote}
    \texttt{batch size} 64, \\
    \texttt{learning rate} $1e-4$, \\
    \texttt{training epochs} 5, \\
\end{quote}

\end{document}